# Deep-XFCT: Deep learning 3D-mineral liberation analysis with micro X-ray fluorescence and computed tomography

Patrick Kin Man Tung[a,b*], Amalia Yunita Halim[a], Huixin Wang[c], Anne Rich[c], Christopher Marjo[c], Klaus Regenauer-Lieb[b,d]

[a]Tyree X-ray CT Facility, Mark Wainwright Analytical Centre, UNSW Sydney, Australia

[b]School of Minerals and Energy Resources Engineering, UNSW Sydney, Australia

[c]Mark Wainwright Analytical Centre, UNSW Sydney, Australia

[d]Western Australian School of Mines, Curtin University, Australia

*Corresponding author: patrick.tung@unsw.edu.au

## Abstract

The rapid development of X-ray micro-computed tomography (µCT) opens new opportunities for 3D analysis of particle and grain-size characterisation, determination of particle densities and shape factors, estimation of mineral associations and liberation and locking. Current practices in mineral liberation analysis are based on 2D representations leading to systematic errors in the extrapolation to volumetric properties. New quantitative methods based on tomographic data are therefore urgently required for characterisation of mineral deposits, mineral processing, characterisation of tailings, rock typing, stratigraphic refinement, reservoir characterisation for applications in the resource industry, environmental and material sciences. To date, no simple non-destructive method exists for 3D mineral liberation analysis. We present a new development based on combining µCT with micro-X-ray fluorescence (µXRF) using deep learning. We demonstrate successful semi-automated multi-modal analysis of a crystalline magmatic rock where the new technique overcomes the difficult task of differentiating feldspar from quartz in µCT data set. The approach is universal and can be extended to any multi-modal and multi-instrument analysis for further refinement. We conclude that the combination of µCT and µXRF already provides a new opportunity for robust 3D mineral liberation analysis in both field and laboratory applications.

## 1 Introduction

To determine the volume fractions of valuable commodities, mineral liberation has become a key step in the mineral processing industry, whereby through the process of breaking up the ore, valuable minerals are released for separation [1]. Mineral liberation analysis is therefore frequently used in metallurgical and mineralogical application for mineral abundance estimation in modal mineralogy and assay techniques for sample elemental distributions [2]. Mineral liberation analysis has also been used in research and industry for particle and grain-size characterisation [3], determination of particle densities and shape factors [4], estimation of mineral associations, liberation and locking [5] and theoretical grade-recovery curves [6]. The technique has now been extended for characterisation of ore deposits to secondary resource characterisation of tailings [7], and applications in the petroleum industry for the characterisation of sediments [8], rock typing, stratigraphic refinement, and reservoir characterisation [9].

Several semi-automated techniques have been developed for identifying and quantifying mineralogy. Of widest use are currently electron-beam based systems using backscattered electron (BSE) images with energy dispersive spectroscopy (EDS) marketed as commercial systems such as QEMSCAN [10], [11]. However, the extrapolation of the collected 2D mineralogy maps to 3D remains problematic. The current methods of stereological reconstruction, at present, cannot solve the difficult mathematical problem of the diversity and complexity of mineral grains in rocks and ores due to the difficult-to-define form factor. This leads to systematic errors in quantitative estimates of volume fractions from 2D data as input to mass-energy balance calculations in mineral processing flow sheets [12]. To rigorously address this problem, 3D tomographic imaging techniques are called for.

Destructive tomographic methods, such as focused ion beam milling coupled with high resolution scanning electron microscope (FIB-SEM) imaging, offer exact 3D reconstructions down to nanometre level and can be complemented by SEM/EDX automated mineralogy [13] on 2D cross-sections. Recent developments in X-ray micro-computed tomography (µCT) have made rapid non-destructive characterisation possible and the technique has now been successfully integrated into the SEM/EDX based mineral liberation analysis [14]. Furthermore, studies have combined micro-X-ray fluorescence (µXRF) with µCT [15] in addition to implementations of artificial intelligence being explored for straightforward segmentations of 2D µXRF maps of rocks [16].

The question arises whether a technique can be successfully developed where the exterior surfaces of an X-ray µCT can be married with a single non-destructive technique for direct, fast, and reliable 3D mineral liberation analysis. The problem is the densities of the geomaterials are so similar that without additional information it is difficult, and often impossible, to reliably differentiate between the distinct phases. This problem is encountered when trying to identify feldspar crystals from crystallized natural magmas where specialised image analysis techniques have been proposed to augment the analysis of tomographic slices through morphological information of feldspar crystals [17]. Though, artificial intelligence, through deep learning protocols, have been found to be useful in overcoming many of the hurdles mentioned in the segmentation of materials [18]–[20], which warrants further development with applications in rocks.

Our hypothesis is the envisioned workflow can be generalised by artificial intelligence techniques that are able to incorporate additional information as well as statistical analysis into a robust 3D mineral liberation analysis technique using laboratory equipment. In this contribution, we propose such a generic multi-instrument –multi-modal analysis technique by combining a deep learning approach and statistical methods for interpreting X-ray µCT and µXRF images, known as Deep X-ray Florescence Computed Tomography (Deep-XFCT). Our aim is to reliably identify and resolve the shapes and distribution of the mineral phases with minor density differences, which will have an effect on mineral flotation as well as the energy needed for rock breakage (comminution); one of the most energy intensive processes in mining [21].

# 2 Methods

## 2.1 Core Plug Rock Sample

### 2.1.1 Multimineral rock sample used in this study

The basaltic andesite core plug sample used in this study was obtained from the NSW Government Londonderry core library. The rock was subdrilled from a 45 mm round rock collection of the Pacific Power Hot Rock 1 bore (PPHR1). The rock was originally from the Muswellbrook geothermal anomaly, located approximately 10 km southwest of Muswellbrook in Hunter Valley Dome Belt, the north-eastern part of the Sydney basin, New South Wales. The PPHR1 borehole was drilled to a depth of 1,946 m near the central region of the anomaly, and more than 1 km of continuous core samples was taken to identify the rocks present and their physical properties [22].

## 2.2 Experimental Methods

To obtain a 3D mineral liberation of the basaltic andesite sample, a combination of conventional analytical techniques was employed for analysis and validation. These included 1) XRD on powder samples of the rock to gain an understanding of the minerals present and their relative abundances, 2) µCT for a 3D representation of the densities of mineral shown in Figure 1, 3) µXRF to obtain the elemental concentration maps from the top and bottom surfaces of the sample shown in Figure 3, and for point analysis of elemental atomic mass, 4) Raman scattering to validate several mineral phases from the current workflow. The XRD and Raman scattering data, along with detailed specifications of the experiment methods can be found in the Supplementary Information.

### 2.2.1 Powder X-Ray Diffraction (XRD)

The XRD was performed to identify the general mineralogy of the rock. A powder specimen was prepared by the backloading method, where random fractions of the rock sample were crushed into powder and sieved to reach a particle size of less than 300 mesh B.S. The powder was dried inside an oven at 50˚C for 24 hours. The data were collected using an Empyrean XRD (PANalytical, Netherlands) fitted with Co as an anode at 45kV and 40mA at the Mark Wainwright Analytical Centre, University of New South Wales (UNSW) Sydney. The collected data were processed by the High Score Plus software (PANalytical, Netherlands). A search match of candidate crystalline phase was performed using PDF 4+ database followed by a Rietveld refinement to quantify the crystalline phases. The data and the Rietveld refinement can be found in the Supplementary Material.

### 2.2.2 Micro Computed Tomography (µCT)

The µCT imaging was performed using the Heliscan µCT system at the Tyree X-ray µCT Facility at the University of New South Wales. The Tyree Heliscan µCT has a General Electric Phoenix Nanofocus tube with a diamond window, a high-quality flatbed detector (3,072 × 3,072 pixels, 3.75 fps readout rate). The facility is built in a lead-lined room with temperature and humidity control (ΔT < 0.5°C). The samples were scanned in helical trajectory with the following setting: 100 kV, 130 µA (tube current), exposure time 0.71s, 5 accumulations, 0.5 mm stainless steel and 0.75 mm aluminium filter, and 2880 projections over a 360º revolution. The voxel size obtained from this sample is 10.79 µm. The tomographic reconstruction was performed using the QMango software. The resulting tomogram of the sample is shown in Figure 1.

### 2.2.3 Micro X-ray Fluorescence (µXRF)

The purpose of µXRF was to provide the mineral labels for the µCT data, and to provide atomic mass percentages of discrete points for mineral verification. The plug was cut into a cylindrical shape to ensure there were two flat surfaces on the top and bottom of the cylinder to obtain µXRF maps from Figure 1. The µXRF mapping was carried out on a benchtop M4 Tornado µXRF (Bruker, Germany) under vacuum condition (20 mbar). The instrument consists of a Rh anode metal-ceramic X-ray tube. A voltage of 45 kV and a current of 600 µA were used. X-ray spot calibration was carried out by digital tuning of the X-ray position indicated on the screen to align with the physical position on the sample. One silicon drift detector is used to count the fluorescent X-rays with an energy resolution ≤145 eV full width at half maximum, for Mn-Kα.

For elemental mapping analysis, a spot size of 20 µm was used and the resulting spectra were acquired every 40 µm with an acquisition time of 200 ms/pixel. Data processing was performed using the M4 Tornado built-in software, FP MQuant (Bruker, Germany) for peak identification and quantification. The data collected from the µXRF maps were XRF intensities which were normalized 8-bit values for all elements analysed. This gave relative elemental concentrations across the whole surface, which provided a semi-quantitative measure (Figure 3). In addition to the µXRF maps obtained for the complete surfaces of the sample, points of interest of size 20 µm were collected (Figure 2) to obtain mass percentages as a further form of quantitative validation.

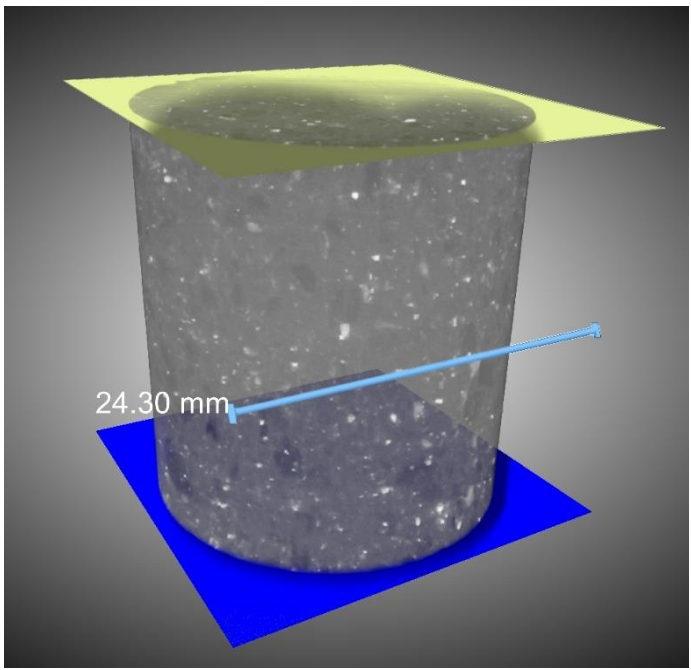

*Figure 1. A µCT dataset of the basaltic andesite sample with the locations of the top and bottom surfaces scanned with µXRF in the yellow (top) and blue (bottom) planes.*

### 2.2.4  Raman Spectroscopy

Specific points were selected based on the mineral phases that were identified from the µXRF maps to confirm the minerals determined through µXRF. Raman spectroscopy was performed using an inVia Raman microspectrometer (Renishaw, UK) using a 532 nm excitation source. Raman spectra were recorded in static mode centred at 1015 $cm^{-1}$ (measurement range: 61-1839 $cm^{-1}$) with a 20x objective, 100% laser power (approximately 33 mW at the sample) with a 1 s exposure and 100 accumulations. Spectra were processed using the WiRE software (Version 5.3, Renishaw, UK) and baseline-corrected to remove the fluorescent background. Spectra were compared to known standards from the online RRUFF database using the phases determined by XRD [23].

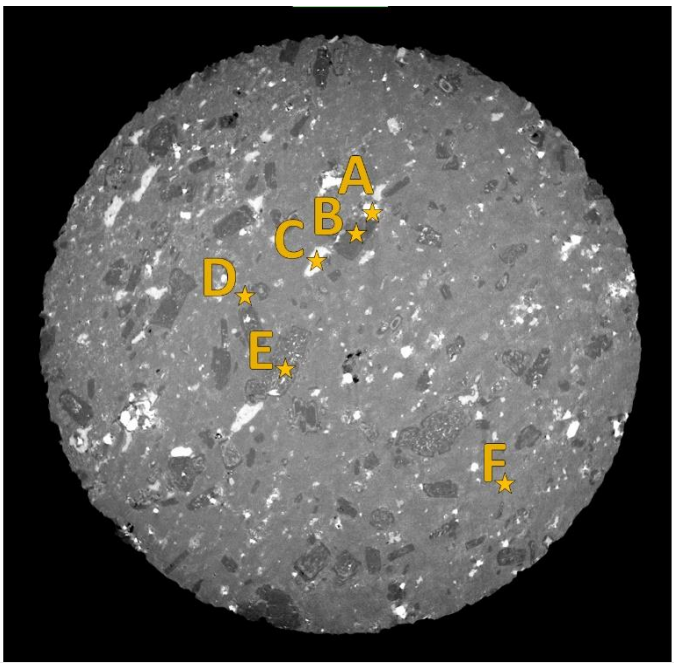

*Figure 2. Positions of point analysis for µXRF and Raman measurements on the bottom surface of the core plug sample. These data was used in extension to the K-means clustering of the µXRF maps and XRD mineral analysis to elucidate mineral segmentation of the surface.*

## 2.3 Data Analytical Methods

The general workflow of the analytical process can be summed in two overarching steps 1) Establish ground truth labelling of mineral phases on the top and bottom of the cylindrical sample through μXRF elemental mapping 2) Use the ground truth labelling of the two surfaces to train a deep learning model that will segment the mineral phases in 3D μCT data. The K-means clustering was performed through Scikit-learn library [24] and the remaining data analytical procedures and visualisations were performed with Dragonfly ORS, including the deep learning segmentation with the Segmentation Wizard [25].

### 2.3.1 Establishing Training Data

To train the deep segmentation learning model to recognise the mineral phases based on the μCT attenuation data, a training data set needs to be established. Here, the training data are 2D images of mineral phases obtained from μXRF μXRF elemental mapping. Not all element maps were used, but the ones used were chosen based on the major elements identified through XRD. For this basaltic andesite sample, the major elements present were Al, Ca, Fe, K, Mg, Mn, Na and Si. The μXRF element maps for the bottom surface are shown in Figure 3, with all remaining maps from all surfaces collected found in the Supplementary Material.

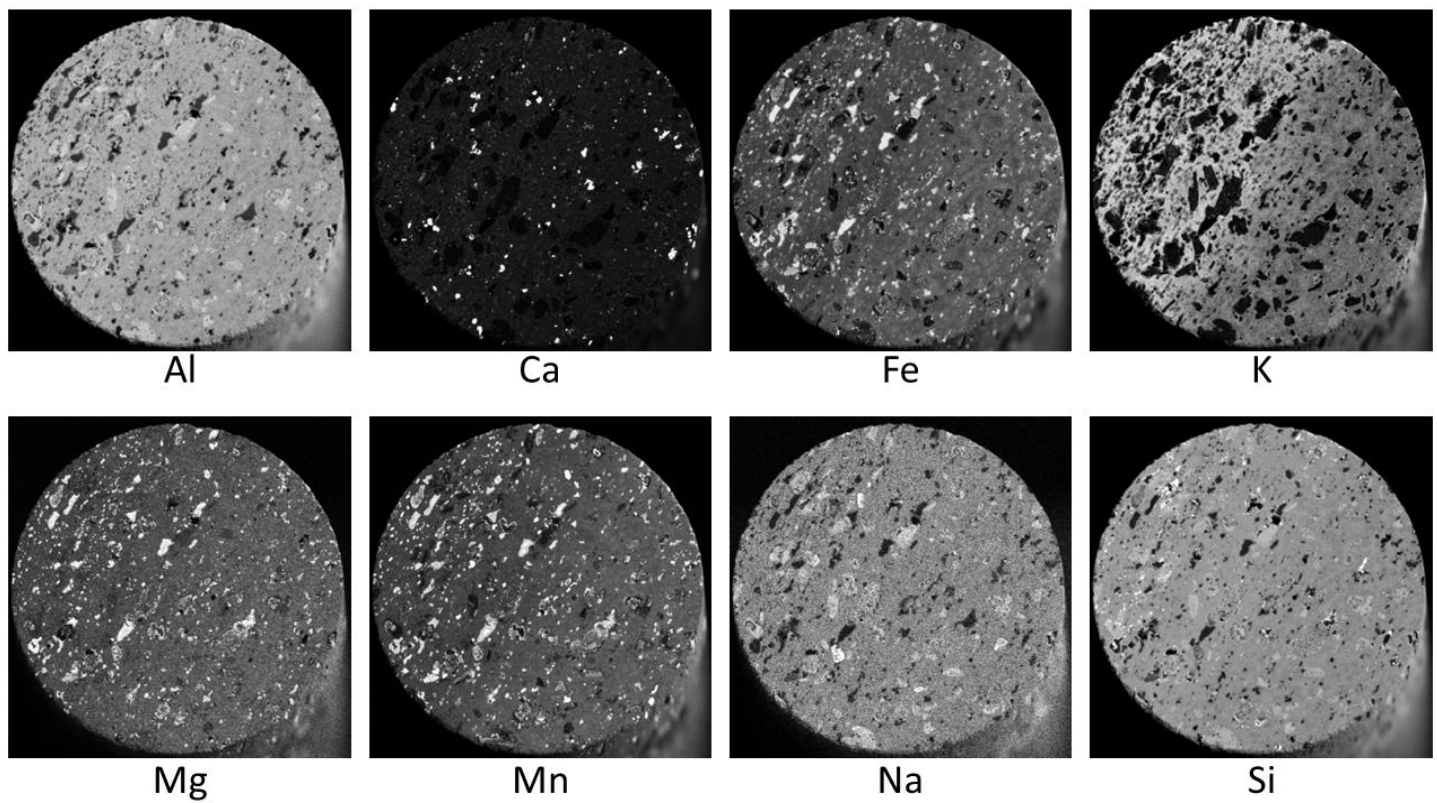

Figure 3. μXRF elemental maps from the bottom surface of the training data shown with greyscale intensities indicating elemental concentrations. The μXRF elemental maps from the remaining surfaces can be found in the Supplementary Material.

#### 2.3.1.1 K-means clustering

K-means clustering was used to automate the identification of mineral phases through the unique information from each elemental map to form the training data for the segmentation model. By grouping XRF concentration intensities from different elements into predefined number of clusters, their clustering centroids serve as a representation of the cluster. This is a fast and simple method that has been shown to extract mineral phases from μXRF maps [16]. Prior to performing the K-means clustering with all μXRF elements, some μXRF maps required pre-processing. This was done for μXRF maps with low atomic numbers such as Na and Mg, where the XRF signals were relatively low and resulted in elemental maps with significant noise, as shown in Figure 4. To address the issue of the noise, various denoising filters can be used. For the specific aim of separating different areas to define minerals, a bilateral filter was used [26]. Bilateral filtering is particularly effective for the application of mineral identification in rocks, as it results in homogenous regions of elemental intensity while maintaining the edges between such regions. The identification of such edges is important for a well-defined dataset to train an effective deep learning segmentation model .

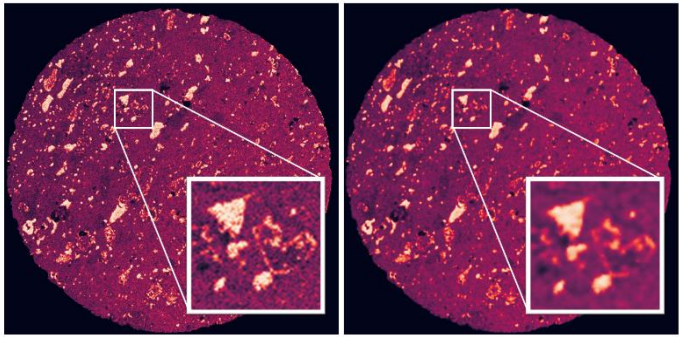

*Figure 4. (Left) Raw µXRF map of Na, where the higher intensities indicate higher concentrations of Na. (Right) µXRF map of Na with bilateral filtering applied. The bilateral filtering reduces noise while preserving the edges, which is important in identifying homogenous grains and defining the interface between grains.*

Seven clusters were used based on the six minerals, recognised from XRD, and the background void. The input data were from µXRF images from Al, Ca, Fe, K, Mg, Mn, Na and Si.  Along with these eight µXRF images, the appropriate orthoslice in the µCT data was also used in the K-means clustering, as shown in Figure 6(a) and (g). This was done to provide added statistical weighting to the greyscale intensities of the µCT images that would be labelled as a result of the clustering. The µCT images were prepared by normalizing the original µCT orthoslice to 8-bit values and resampling the µCT image with approximately 10 µm pixels into the geometry of the µXRF images with 40 µm pixels.

Following the pre-processing of the Na and Mg µXRF maps, K-means clustering was applied to segment minerals by combining all eight relevant elemental maps, with each elemental map serving as a dimension in the clustering space. The clustering was performed through the Scikit-learn Python library [24], where the best result from 10 random initial clustering centroids was taken. The final centroids output represented the minerals and were used as labels for training the deep learning segmentation model (Table 2).

### 2.3.1.2  Image Registration

To perform the registration of the 2D mineral image from the K-means clustering analysis to the X-ray µCT volume, a µXRF image with high contrast and similar structures evident was used. In this case, the Fe elemental µXRF maps were used. A coarse registration was first performed with manual manipulation of the µXRF within the µCT volume, where the mobile data was the µCT volume as it is higher resolution and in 3 dimensions. Then an automatic image registration was performed for finer alignment. The automatic image registration algorithm used was the mutual information registration method with linear interpolation [27], as shown in Figure 5.

Once the µXRF image is registered within the µCT volume, the voxel orientations of the µCT volume was aligned with the µXRF map. This allowed the representative orthoslice from the µCT volume to be extracted and the µXRF map to be resampled into the voxel geometry of the appropriate µCT orthoslice. Finally, the relevant pixels of the µCT sample were used to crop the µXRF map so that XRF signals from the edge of the sample that did not represent the sample would be removed. This results in a mineral label for each µCT pixel in the orthoslice.

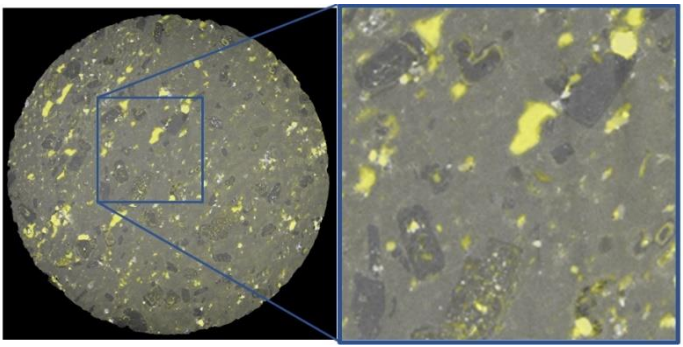

*Figure 5. An opaque overlay of a Fe µXRF map (yellow) over the matching µCT orthoslice showing minor inaccuracies in the registration between the two datasets.*

### 2.3.2 Deep learning segmentation

The training data for the deep learning model was created from µXRF images that were registered and resampled into the µCT geometry so that each pixel in the µCT image had a defined label from the µXRF images. A total of 80% of the labelled data was used for training the model, and the remaining 20% was used for validating the model. The training data was divided into patches of 128 x 128 pixels. Data augmentation was then applied to add an extra 10 sets of data. This included flipping the patches vertically and horizontally, rotations of 90º or 180º, zooming of 0.9 to 1.1 and a shearing factor of 2.

The model architecture used was U-net with a depth level of 5 layers [28]. A validation loss function of ORS Dice Loss was used to assess the performance of the segmentation model [25]. The model was trained with 200 epochs with a batch size of 16, stride ratio of 1.0 with an initial learning rate of 1.0 with the Adadelta optimization algorithm [29] that reduced by 50% when the loss function did not decrease for 10 consecutive epochs.

Due to the concern of limited training data from the top and bottom of the sample, the criterion for choosing the best segmentation was not only filtered by the minimum validation loss, but also through visual analysis. This was done by saving the 5 best models based on the lowest validation loss and selecting the best model based on user-visual inspection. The final segmentation model selected had a loss value of 0.6749.

The deep learning segmentation was performed in Dragonfly ORS software version 2021.2 [25].The total training time was 191 minutes using an Intel Core i9 10900K central processing unit and a NVIDIA RTX 3090 graphics processing unit.

### 2.3.3 Establishing Validation Data

While the aim of the workflow is to use the top and bottom surface of the cylindrical plug to extract the mineral phases throughout the volume, a true test of the generalisability of the model can only be done by comparing with other surfaces not exposed to the deep learning model. Hence, to obtain such a ground truth dataset, µXRF maps were also obtained from surfaces from within the bulk by cutting and removing a section of the sample. The resulting surfaces can be seen in Figure 8. This allowed the performance of the segmentation model to be compared to µXRF data it was not exposed to.

The same K-means clustering, and image registration procedures used to make the training data were also used to produce the ground truth data.

# 3 Results

## 3.1 Multi-modal mineral assignment with XRD and XRF

### 3.1.1 Multi-modal data

For the deep learning training data, each pixel of the image must be labelled with an identified mineral. To achieve this, XRD, and XRF data were combined to identify the minerals at each pixel with Raman data to reinforce mineral assignments. For ease of reference, the tables below show the information used from XRD and µXRF mapping K-means clustering and point analysis from XRF and Raman, respectively.

*Table 1. XRD results: mineral phases identified and their respective abundance percentages from a random selection of the sample.*

| Mineral Identified | Chemical Formula | Abundance (%) |
|---|---|---|
| Albite | $Al_{1.488}Ca_{0.491}Na_{0.499}O_8Si_{2.506}$ | 50.4 (±5.6) |
| Ankerite | $C_2Ca_{0.997}Fe_{0.676}Mg_{0.273}Mn_{0.054}O_6$ | 1.4 (±0.1) |
| Clinochlore | $H_{16}Al_{2.884}Fe_{0.874}Mg_{11.126}O_{36}Si_{5.116}$ | 9.7 (±0.8) |
| Illite 2M | $C_2Al_4K_1O_{12}Si_{12}$ | 12.0 (±1.5) |
| Laumontite | $H_4Al_2Ca_1O_{14}Si_4$ | 14.0 (±1.9) |
| Quartz | $SiO_2$ | 12.6 (±0.9) |

*Table 2. µXRF K-means clustering results: clustering centroids for µXRF maps of the bottom surface shown as normalised 8-bit fluorescence intensities, with a minimum of 0 and maximum of 255. Additionally, the abundance percentage of the labels are shown, along with the assigned mineral.*

| Assigned Mineral | Centroid number | Colour on images | Al (Ct.) | Ca (Ct.) | Fe (Ct.) | K (Ct.) | Mg (Ct.) | Mn (Ct.) | Na (Ct.) | Si (Ct.) | µCT (Ct.) | Abundance (%) |
|---|---|---|---|---|---|---|---|---|---|---|---|---|
| Albite | 1 | Red | 149 | 18 | 91 | 105 | 89 | 86 | 134 | 148 | 156 | 48 |
| Ankerite | 2 | Grey | 45 | 134 | 126 | 31 | 65 | 92 | 66 | 53 | 203 | 1.98 |
| Clinochlore | 3 | Yellow | 72 | 13 | 191 | 20 | 203 | 220 | 67 | 73 | 187 | 4.06 |
| Illite | 4 | Light blue | 163 | 16 | 74 | 167 | 81 | 75 | 136 | 159 | 150 | 24.8 |
| Laumontite | 5 | White | 121 | 16 | 118 | 44 | 131 | 139 | 113 | 132 | 155 | 7.06 |
| Quartz | 6 | Dark blue | 182 | 12 | 36 | 27 | 72 | 52 | 182 | 179 | 120 | 14 |

*Table 3. Point analysis results from XRF and Raman: Mass percent from XRF and detection from Raman of selected 20 µm spots in Figure 2. Note, for the XRF mass percentages, other trace elements detected were ignored as they were not in the minerals detected by XRD.*

| Assigned Mineral | Label in Figure 2 | Raman detection | Al mass % | Ca mass % | Fe mass % | K mass % | Mg mass % | Mn mass % | Na mass % | Si mass % |
|---|---|---|---|---|---|---|---|---|---|---|
| Albite | F | Detected | 16.60 | 2.61 | 11.93 | 4.88 | 1.66 | 0.24 | 8.82 | 51.82 |
| Ankerite | A | Undetected | 16.12 | 3.23 | 18.39 | 3.57 | 4.69 | 0.56 | 6.33 | 45.54 |
| Clinochlore | C | Detected | 15.88 | 3.13 | 13.23 | 4.94 | 2.59 | 0.33 | 7.24 | 51.10 |
| Illite | D | Undetected | 16.59 | 3.14 | 10.30 | 6.77 | 1.09 | 0.23 | 7.61 | 52.74 |
| Laumontite | E | Undetected | 16.29 | 3.28 | 10.33 | 6.41 | 1.42 | 0.22 | 7.70 | 52.73 |
| Quartz | B | Detected | 16.35 | 2.03 | 8.72 | 2.37 | 1.96 | 0.24 | 8.79 | 58.64 |

### 3.1.2 Mineral assignment

For this basaltic andesite sample, the phase identification can be performed entirely through the use of the XRD and µXRF data. The µXRF data includes the normalized concentration maps for each element and the mass percentages of spots from point analysis. Further validation of minerals was obtained through Raman spectroscopy.

Initially, the information from XRD, shown in Table 1, was used to identify the mineral phases that occur throughout the bulk of the sample. The XRD analysis provided three prerequisite details of information. First, the mineral phases that exist within the sample that holds critical elemental information to correlate with the µXRF data. Second, the number of mineral phases that was obtained, which guided the number of clusters for the K-means clustering. Third, the abundance percentages of mineral phases throughout the sample, which served as a form of validation for cross-referencing with the K-means clustering abundances and the accuracy of the segmentation from deep learning. The XRD pattern and the Rietveld fitting [30] for mineral identification and quantification can be found in the Supplementary Material.

The centroids for the K-means clustering that represent each mineral present at the bottom surface of the sample are listed in Table 2, and the resulting mineral segmentation is shown in Figure 6(b) and (h). The assigned minerals from the K-means clustering are also listed in Table 2, however, it is not possible to determine the minerals based solely from the K-means centroids. This is because the XRF intensities represent the concentration of elements at the analysis point and do not correlate directly with the chemical formulas of discrete minerals. Therefore, the centroids give an indication of elementally rich or low areas, which can only guide mineral identification. Instead, this information is merged and extended with elemental mass percentages from XRF point analysis to obtain mineral assignments. Points were selected on areas that had large homogenous regions to ensure the correct representation of the mineral.

By applying geochemical principles to the K-means clustering centroids of the µXRF maps, shown in Table 2, there are three mineral assignments that can be made: 1) First, illite can be unambiguously identified based on the minerals identified in the XRD, as illite is the only mineral that contains potassium. Hence, the $4^{th}$ K-means centroid with the highest intensity of potassium can be associated with illite. Further, the centroid also has relatively high concentrations of Fe and Si. 2) Second, ankerite can be identified with the $2^{nd}$ centroid with the highest Ca concentration, which also contains high concentrations of Fe. Additionally, the abundance also correlates with the K-means clustering and the XRD at approximately 2% each. 3) Third, the $5^{th}$ centroid with the highest concentration of Mg can be used to identify clinochlore as that is the remaining mineral with Mg, as ankerite has already been identified. Moreover, clinochlore is the only mineral with high concentrations of Mg and Fe. This identification of the cluster with clinochlore has also been confirmed through Raman scattering.

From the remaining K-means centroid concentrations, there are no unique differentiators to identify the remaining minerals of albite, laumontite and quartz. The remaining mineral identification can be performed with the addition of mass percentages from XRF point analysis, shown in Table 3. 4) Fourth, quartz can be identified by the point with the highest Si mass percentage. This is further evidenced by the majority of other points having low Si mass percentages. Furthermore, this mineral is confirmed through Raman. 5) Fifth, albite can be identified with the use of the highest Na mass percentage as it is the only mineral with Na. Additionally, both the K-means clustering and XRD indicate this mineral phase to be the most abundant. Furthermore, this mineral is confirmed through Raman. 6) Finally, laumontite did not have strong indications from the µXRF data, as there was no unique elemental identifier or characteristic combination of elements. Though, through deduction as the final mineral, laumontite can be identified.

Hence, the phase identification can be performed on this basaltic andesite sample. This can be performed by using combination of mineral information from XRD, the intensity centroids from µXRF maps, their relative abundance, and the mass percentages from point analysis. However, the demonstration shows that by using the information from XRF exclusively, one could extract the mineral phases. Further, if select locations can be phase-identified through Raman analysis, then phase identification would be straightforward.

## 3.2 Deep Learning Segmentation on µCT data

### 3.2.1 Training data

With minerals assigned to each pixel of the µCT orthoslice on the top and the bottom surface of the sample, this information can be propagated through the 3D tomogram through a deep learning segmentation model. The grayscale µCT data of the top and bottom slices used are shown in Figure 6(a) and (g). The labelled mineral phases of the slices were then used for training the segmentation model as shown in Figure 6(b) and (h). The resulting segmentation from the trained model are shown in Figure 6(c) and (i). Zoomed in sections for each respective image are shown in Figure 6(d), (e), (f), (j), (k), (l). From a visual analysis, the model has captured the major contours of the mineral phases. This includes separating the abundant matrix minerals of albite and illite to a good approximation, despite the greyscale intensities appearing identical to the eye. However, the model fails to recognise the finer details apparent in the labelled training data. The overall accuracy from all the mineral labels for the top and bottom surface were 70.23% and 69.56% with the voids excluded, respectively.

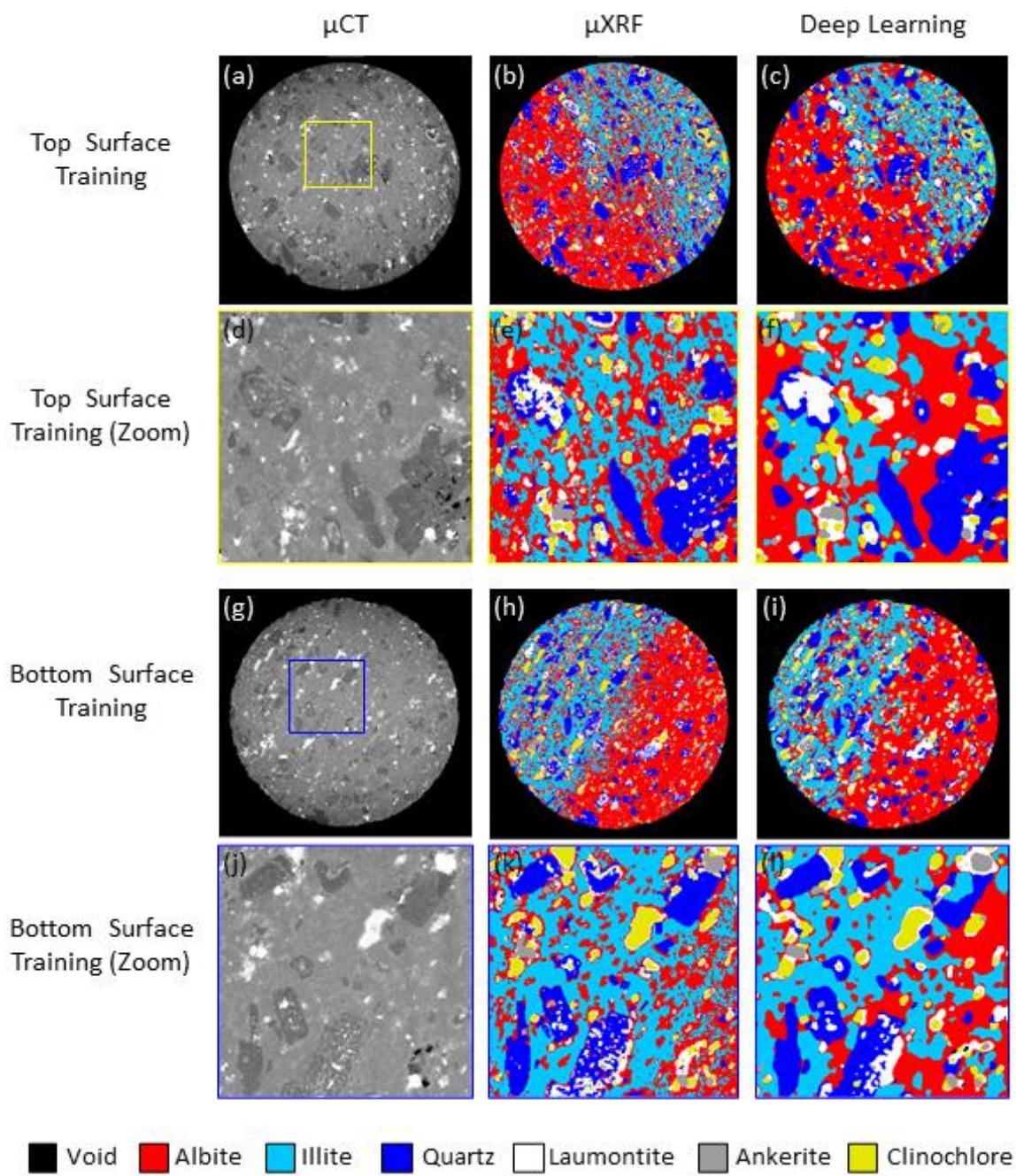

*Figure 6. Training data from the top surface for (a) μCT data, (b) μXRF from K-means clustering and (c) deep learning segmentation with (d), (e) and (f) showing zoomed areas as defined by the yellow inset box, respectively. The training data from the bottom surface is shown for (g) μCT data, (h) μXRF from K-means clustering and (i) deep learning segmentation with (j), (k) and (l) showing zoomed areas as defined by the blue inset box, respectively.*

Further analysis of the pixel accuracy of the segmentation model is represented as a confusion matrix, as shown in Figure 7. The confusion matrix shows the proportion of predicted mineral pixel labels from the segmentation model for each mineral class with the true mineral pixel labels from the μXRF maps. This provides insights into the types of errors made by the model – in essence what predictions the model is confused by. For a model with high accuracy, the confusion matrix would possess high accuracies along the diagonal, from top left to bottom right. The overall profile of the confusion matrix between the two surfaces measured is similar, with similar pixel accuracies across all mineral phases. This indicates the consistency of the segmentation performance on different surfaces.

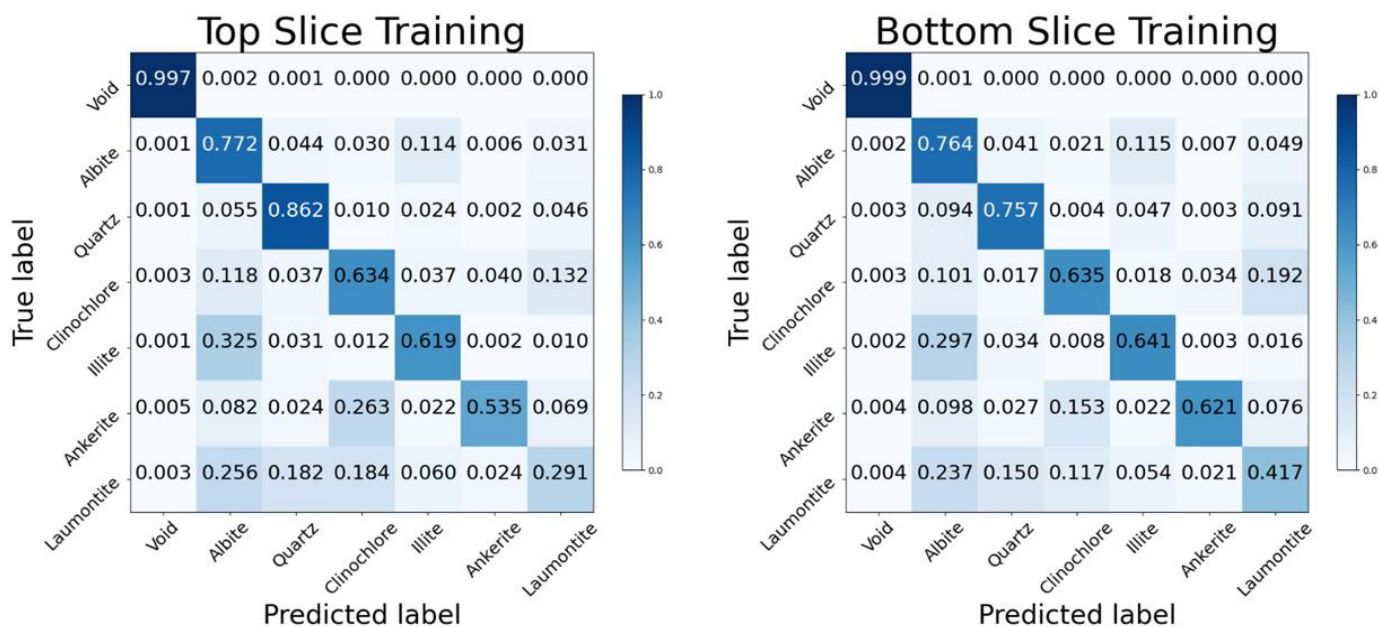

*Figure 7. Pixel-wise confusion matrix for the segmentation of the top and bottom surface, showing the ratio of correctly and incorrectly predicted labels of the segmentation model in comparison to the true labels from the μXRF maps.*

The accuracy of the void phase occurring mainly outside of the sample was correctly identified and hence the model was able to discern from sample and empty space. Meanwhile, the mineral phases within the material varied in accuracy. The two predominant minerals consisted of albite and illite, as shown in Figure 6. The model was able to discern the difference between the two primary mineral phases even though the phases are visually difficult to distinguish due to the similar greyscale intensities. Additionally, the model was able to overcome the intensity gradient from beam hardening to match the general segmentation from the training data. However, as can be seen in the confusion matrix for both surfaces in Figure 7, the majority of the incorrect labels for each of the two phases were attributed to the other phase. This suggests the task is inherently challenging and the model cannot separate the albite and illite confidently.

For the discrete grains of minerals, the quartz phase is identified with the highest accuracy from all minerals. Many of its false labels were from the surrounding minerals of albite and illite, the other major contributor is laumontite, which typically occurs in and around the quartz grains, and this is to be expected. Laumontite is the least accurate phase with less than half the pixels labelled correctly for both surfaces, where the low quality of segmentation is also apparent from visual inspection. This is because the occurrence of laumontite is often intricate fine structures, which requires highly accurate registered μXRF data to provide the model with strongly relevant training data. This has resulted in mislabelling with mostly albite, quartz and clinochlore. For clinochlore, 63% of the pixels were labelled correctly and mislabelled pixels were mainly laumontite and albite. Finally, ankerite had an average of 58% of pixels correctly identified with the largest proportion of mislabelling occurring with clinochlore. This is because both ankerite and clinochlore are relatively dense causing a similar greyscale tomography intensity. The accuracy is further compounded by the relatively small abundance of ankerite whereby any mislabelling would constitute a large percentage of labels.

### 3.2.2 Validation data

While the model performed well on segmenting the training data, the true test for generalisability is to validate how well the model performs for data it has not been exposed to at all. In this case, this would be additional labelled slices within the sample. For this purpose, a thin slice (ca. 2mm) of the plug was removed from the bottom and top part of the sample to expose such surfaces to generate validation data.

Figure 8(b) and (h) show the mineral labels for the top and bottoms surfaces from within the bulk of the sample. The same mineral identification approach was performed as the trained data set by using K-means clustering analysis of µXRF maps. Figure 8(c) and (i) show the segmentation results from the model that was trained from the outer slices. From visual inspection, the segmentation model performs well in extracting the defined course-grained minerals. In comparison to the training data, the model was able to recognise the approximate distribution of the matrix minerals of albite and illite, though not to the same accurate degree as the training data in Figure 6. In general, the performance on the validation data is lower than the training data, as shown with accuracies of 63.50% and 62.79% from the top and bottom surface with the voids excluded, respectively.

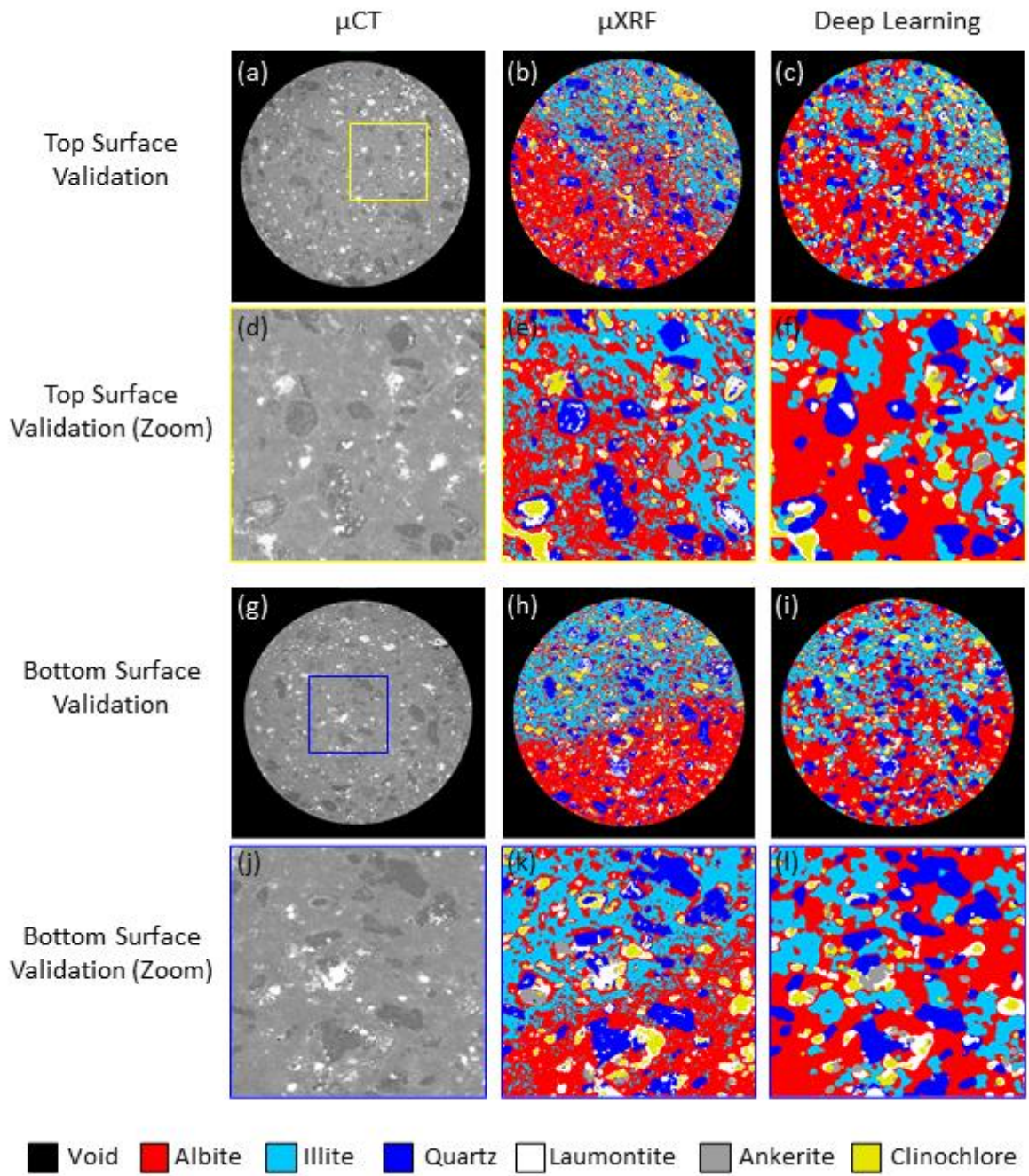

*Figure 8. Validation data from the top surface for (a) µCT data, (b) µXRF from K-means clustering and (c) deep learning segmentation with (d), (e) and (f) showing zoomed areas as defined by the yellow inset box, respectively. The validation data from bottom surface is shown for (g) µCT data, (h) µXRF from K-means clustering and (i) deep learning segmentation with (j), (k) and (l) showing zoomed areas as defined by the blue inset box, respectively.*

Figure 9 shows the confusion matrix for the validation data. As with the training surfaces, the profile of the confusion matrix is similar with the validation surfaces, with a comparable spread of correct and incorrect label accuracies. Based on the two predominant minerals of albite and illite, there is an overprediction of the albite labels and underprediction of the illite labels. Quartz conversely is predicted well by the model with over 82-84% of the phase being predicted correctly. Clinochlore has an accuracy of 56-61% for both slices. Similar to the segmentation on the training data, the mislabels are mainly attributed to laumontite and, to a lesser extent, albite. Meanwhile, laumontite reaches an accuracy of 26-28% for both slices. The mislabels occur significantly across all other mineral phase labels. As previously suggested, this may be caused by a combination of the inappropriately minimal registration of the indistinct structure and the ambiguous greyscale intensity in the µCT dataset. Finally, ankerite has different performances across the two slices. For the top slice, the accuracy reaches 44% while for the bottom slice it is 38%. The majority of the mislabels occur with clinochlore. This is due to having similar densities which results in similar greyscale values in the µCT dataset.

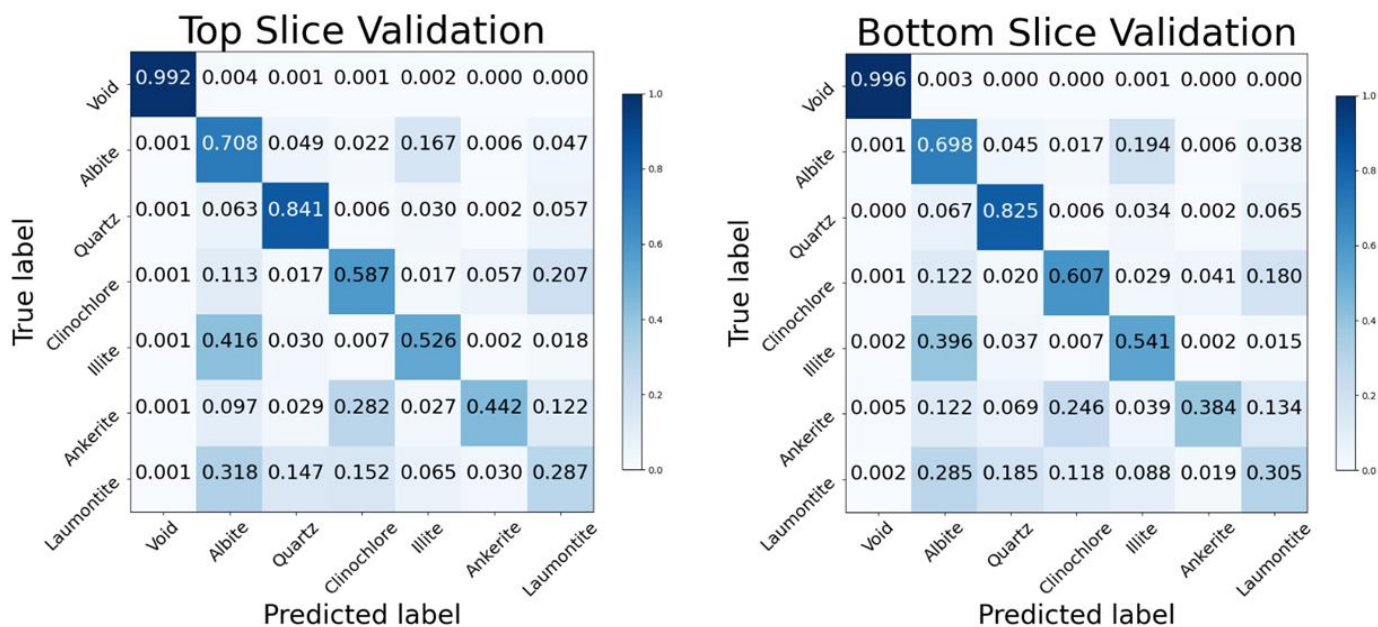

*Figure 9. Confusion matrix showing the predicted deep learning and true µXRF mineral pixel classifications for the validation data collected from surfaces within the bulk of the basaltic andesite rock sample.*

Figure 10 shows the 3D volume rendering of the tomography collected from the basaltic andesite sample in addition to the deep learning segmentation that was trained on the top and bottom surface of the sample. In terms of segmentation quality, the coarse grains show clear inconsistencies in the segmentation of the 2D orthoslices, although the model can distinctly identify the position and shape of the grains. Moreover, the model was also able to generalise details of the coarse-grained mineral structures. For example, the higher density inclusions within and around quartz were specifically recognised as laumontite, while other similarly dense minerals were correctly identified as clinochlore. For the fine-grained minerals of albite and illite, the model showed the ability to separate these minerals, which have faint intensity differences that are difficult to separate from visual observation and conventional thresholding methods. With the perspective of the 3D volume rendering, it is also possible to see apparent striations of the minerals perpendicular to the long axis of the cylindrical sample. This is because the segmentation model was 2-dimensional and was applied on each orthoslice. Hence, this shows a lack of connection in terms of contextual segmentation information between the orthoslices and limits the continuity along the long axis of the cylindrical sample.

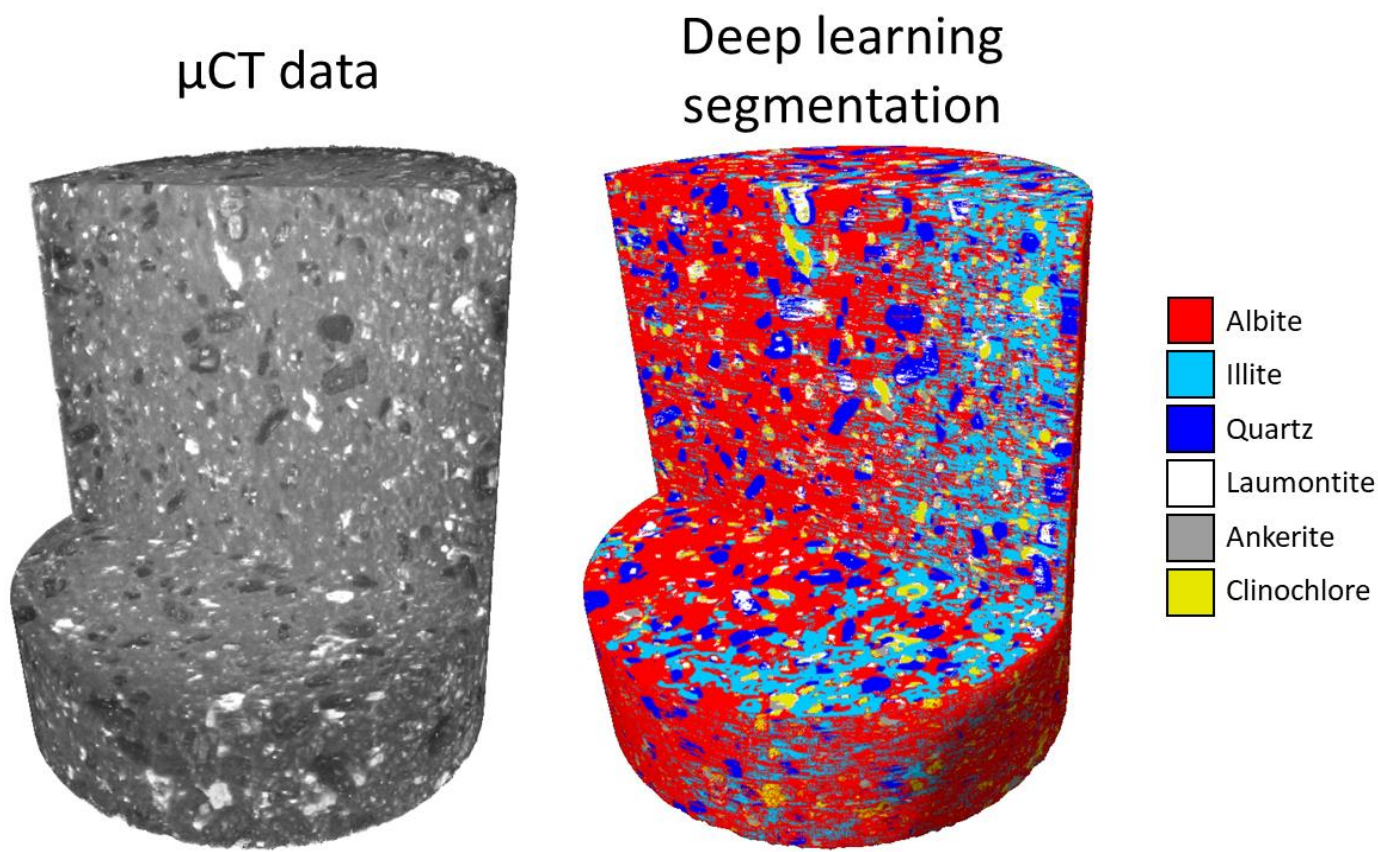

*Figure 10. 3D tomography cutaway of the cylindrical basaltic andesite core plug (left) and the 3D segmentation performed by the deep learning model (right).*

The relative abundances of the minerals throughout the sample from the Deep-XFCT segmentation is shown in Table 4. The abundances are in good agreement with the results obtained from µXRF mineral analysis of the top and bottom surface of the sample, shown in Table 2 and Table 3 respectively. Evidently, albite comprises the majority of the sample at approximately half of the sample, followed by illite, quartz, laumontite, clincochlore and ankerite. However, these mineral abundances in comparison with XRD (Table 1) show some discrepancies, specifically showing a lower illite and higher laumontite and clinochlore abundance. These discrepancies are within the margin of error when considering that the smaller volume sampled by XRD may not be representative of the whole core plug, as can be seen by the mineral distribution heterogeneity. Given XRD is an important tool to validate the mineral phases, it is important to try and ensure sampling is representative of the entire core.

*Table 4. Abundance percentage of each mineral obtained from segmentation with the deep learning model.*

| Mineral | Deep-XFCT segmentation abundance (%) |
|---|---|
| Albite | 50.48 |
| Ankerite | 1.85 |
| Clinochlore | 5.18 |
| Illite | 20.71 |
| Laumontite | 7.06 |
| Quartz | 14.71 |

# 4 Discussion

In comparison to existing techniques that probe similar mineral domains, Deep-XFCT introduces a novel approach that exploits the benefits of artificial intelligence. Specifically, Deep-XFCT is able to extract a fully segmented 3D grain structure non-destructively and at large enough scales to accommodate a cylindrical core plug with measurements of approximately 24.3 mm in diameter and 25.9 mm in length. The results show Deep-XFCT works well in practice and should compare favourably to the conventional approaches of segmenting minerals, such as intensity thresholding, top-hat or watershed segmentation that can be affected by intensity inhomogeneity from beam hardening and experimental artefacts. In the context of its intended purpose for 3D mineral liberation analysis, Deep-XFCT can isolate grain locations and provide clear indications and realistic perspectives of the 3D morphology, distribution and abundance of minerals throughout the bulk volume of the core plug.

Furthermore, the use of deep learning extends the possible capabilities of segmentation through computer vision and removes the significant contribution of user bias to provide an objective approach with mineral segmentation. This is exemplified with the model to not only identify the coarse grains, but to also to discern minerals of similar greyscale intensities between the fine-grained minerals of albite and illite. Such a result would be valuable for mineral liberation analysis and unattainable through conventional methods. It could be argued that the distribution of albite and illite are similar throughout the length of the core plug sample, and therefore the model generalised the distribution. However, the model works in smaller isolated patches of each orthoslice, which would mean the model can recognise the imperceptible differences in intensity to assign either albite or illite. Furthermore, Deep-XFCT possesses advantages of minimal sample preparation and uses relatively accessible instrumentation and software to provide an approachable and practical method of 3D mineral liberation analysis.

### 4.1.1 Contributing factors to results

In comparison to existing methods that provide similar mineral information, such as QEMSCAN and FIB-SEM tomography, Deep-XFCT extends past surface information into large volumes and in a non-destructive manner. However, the two techniques have several advantages over the Deep-XFCT approach. QEMSCAN and FIB-SEM tomography can extract mineral information at higher resolutions due to the inherent properties of electrons in comparison to X-rays, although this can result in sampling bias if only small areas of the sample are imaged. Further, with Deep-XFCT, there is more manual handling of the data at this point of development in contrast to the QEMSCAN and FIB-SEM tomography. Moreover, there are several universal factors that can contribute to the inaccurate mislabelling of minerals for the basaltic andesite sample that limit the performance of the current implementation of Deep-XFCT. Though at the core of the problem all such issues stem from the mismatch of the µXRF and µCT data to produce accurate training data with correct pixel labels for the deep learning model, which will be discussed in the following.

Firstly, the registration between the µXRF and µCT data is not perfectly aligned, as there is not a perfect correspondence between the grain shapes between the two datasets (Figure 5). Therefore, at a per-pixel scale, the information from µCT likely does not correlate completely with the mineral phase identification, leading to unambiguously incorrect labels for the µCT data. Further, defining the interfaces between grains is important for the network to recognise the features. This issue is exemplified by laumontite as it is commonly appears in fine intricate structures in and around quartz and has shown to be difficult to register perfectly. The addition of an added degree of freedom from the 2D orthoslice to the 3D tomogram also introduces imperfections in the registration. However, at the larger scale of the coarse mineral grains, the imperfect registration does not affect the overall performance as there is a representative quantity of correct labels to overcome the incorrect labels.

Secondly, the resolution of both the µXRF and µCT measurements limits the accuracy of the resulting segmentation model. This was the case with the µXRF data, where the pixel resolution was 40 µm as opposed to the approximately 10 µm from the µCT data. The up-sampling of the µXRF pixels into the higher resolution geometry of the µCT pixels would lead to incorrect labelling of the µCT pixels, which would cause the model to have invariably incorrectly labelled training data. Therefore, the limiting resolution of the µXRF data also prohibits the method from being implemented on fine-grained minerals for the time being and restricted to coarse-grained minerals. However, based on the µXRF instrumentation, the resolution can be improved to 10 microns on smaller areas with the drawback of reducing the quantity of training data.

The issues mentioned relating to mismatching datasets are unavoidable when using inherently different techniques and must be viewed in their usefulness for the desired applications. A possible approach to resolve many of the issues would be to introduce an intermediary step within the segmentation workflow. This would involve first training a traditional machine learning model with a sparse dataset that does not require all pixels in the training data to be labelled. The training data could be produced through an erosion process of the defined grains from the µXRF analysis. This machine learning model would recognise minerals by their greyscale intensity in the µCT data and provide a segmentation that would define the interfaces better than the raw µXRF input. As a result, this segmentation from a traditional machine learning model could be more appropriate as a dense training set with all pixels labelled for training the deep learning model.

In terms of discussing the aspects of the deep learning process, an obvious concern is whether the quantity of training data is adequate, as having more training data is usually desirable. The training data sourced from the top and bottom of the sample appears to be sufficient for training the model to recognise the features required for mineral liberation analysis. This is contingent on the training data having satisfactory representation of all minerals, interfaces, shapes and features that would be present throughout the sample. The quantity of training data could be increased by cutting the cylindrical core plug into a prism with additional flat surfaces suitable for µXRF analysis. Another concern relates to the relatively large patch size of 128 x 128 pixels that may have affected results; where the larger the patch size, the greater loss in detecting smaller details [31]. Conversely, a smaller patch size would not be able to access the larger features of minerals outside of each patch, limiting its usefulness. Therefore, the patch size selected for the deep learning model, whilst not able to discern the finer mineralogical details, was the best compromise in balancing segmented details at different scales.

Considering the possible limitations and shortcomings of the Deep-XFCT technique, the overarching goal to extract minerals for the purposes of mineral liberation analysis was successfully performed for this specific collection of minerals.

### 4.1.2  Segmentation enhancement pathways

While the Deep-XFCT approach performs well, there are multiple pathways of improvement that can be implemented. All improvements should focus on the main objective of obtaining the most accurate segmentation possible. To this end, the improvements can be separated into those for preparing and obtaining the best training data and those related to training the model.

The most simple and obvious improvement for the training data is increasing the quantity of training data. This can be achieved by increasing flat surfaces on the sample by trimming off sections, for example into a prism shape, to obtain more µXRF data. However, this may not be practical or preferable as the sample may need to be kept intact and more time is required to collect additional µXRF data. Similarly, the quality of the training is also important, and this can be facilitated through higher resolutions datasets from both µCT and µXRF by reducing the overall sample size to provide more detailed and accurate training data. Furthermore, to increase the statistical robustness of the clustering method on the µXRF maps, a high-resolution photo can be taken to be added as part of the K-means clustering data set to provide another dimension of information in the form of visible light information. Finally, while the finer registration is automated, the coarse registration of finding the corresponding orthoslice within the 3D tomography data is manually intensive and tedious. The coarse registration can also be optimised through artificial intelligence, allowing the whole Deep-XFCT procedure to be streamlined through automation [32].

For the process of training the segmentation model, a possible enhancement to the original U-net model is to use a multiscale U-net model instead [33]. The architecture of the multi-scale patch-based model would use multiple patch sizes to capture different contextual information at various dimension scales. This is done by creating a model input image by separating the initial image into overlapping patches, followed by combining with smaller up-sampled patches and larger down-sampled patches. This would accommodate different dimension scales that will allow the elucidation of all mineral mineralogy within the sample. Additionally, the use of a 3D U-net model [34] would be able to remove the striation artifacts seen in the resulting segmentation when using a 2D model. However, the need for 3D training data for a 3D segmentation model would be difficult to manifest from µXRF maps collected from surfaces. Further, there is a significant increase in computation cost to using a 3D segmentation model over a 2D model and must be considered based on the results required.

# 5 Conclusion

Deep-XFCT uses deep learning as the correlative bridge between the different modalities and dimensions of µXRF and µCT. A deep learning segmentation model was trained with the minerals identified on the top and bottom surface of a cylindrical core plug and segmentation results were validated with surfaces from within the original core plug. While Deep-XFCT does not provide pixel-level accurate segmentations, it has been demonstrated to provide an excellent proxy for the purposes of obtaining the presence, location, distribution, and morphology of grains dispersed throughout a core plug. Furthermore, this artificial intelligence augmented workflow was shown to segment fine-grained minerals of similar density that would be impossible by manual segmentation. Moreover, it is also a conceptually simple workflow that uses experimental instrumentation that is readily accessible at research and commercial institutions. It also possesses all the advantages of no sample preparation required, analysis of large samples of full core plugs and is non-destructive. Hence, Deep-XFCT is a viable and valuable technique that has the potential to expedite 3D mineral liberation analysis.

While the technique has been demonstrated and optimised on a geological core plug example for mineral liberation applications, the fundamental workflow of combining unique information from the surface and propagating through a volume via deep learning algorithms can be adopted by a plethora of image-based techniques. With such vast directions in which the technique can grow and be applied, the current approach is foreseen to be a pivotal steppingstone in incorporating various modalities of analysis into comprehensible and enriching findings through artificial intelligence.

# 6 Acknowledgements

The authors would like to acknowledge Simone Zanoni from the Biological Resources Imaging Laboratory, UNSW Sydney for providing background information on image analysis and registration. The authors acknowledge the use of the facilities at the Mark Wainwright Analytical Centre, UNSW Sydney, including the Spectroscopy Laboratory, X-ray Fluorescence Laboratory, X-ray Diffraction Laboratory and the Tyree X-ray CT Facility. K.R. would like to acknowledge support from the Australian Research Council (ARC DP170104550, DP170104557, LP170100233 and LE200100209). P.K.M.T. would like to acknowledge support from an internal UNSW grant RIS RG193860 'Multi-instrument, multi-modal co registration of tomographic images.'

# 7 Conflict of Interest

The authors do not have any conflicts of interest.

# 8  Data Availability

The raw and processed data required to reproduce these findings cannot be shared at this time due to technical or time limitations. The raw and processed data can be obtained from the corresponding author upon reasonable request.

# 10 Supplementary Material

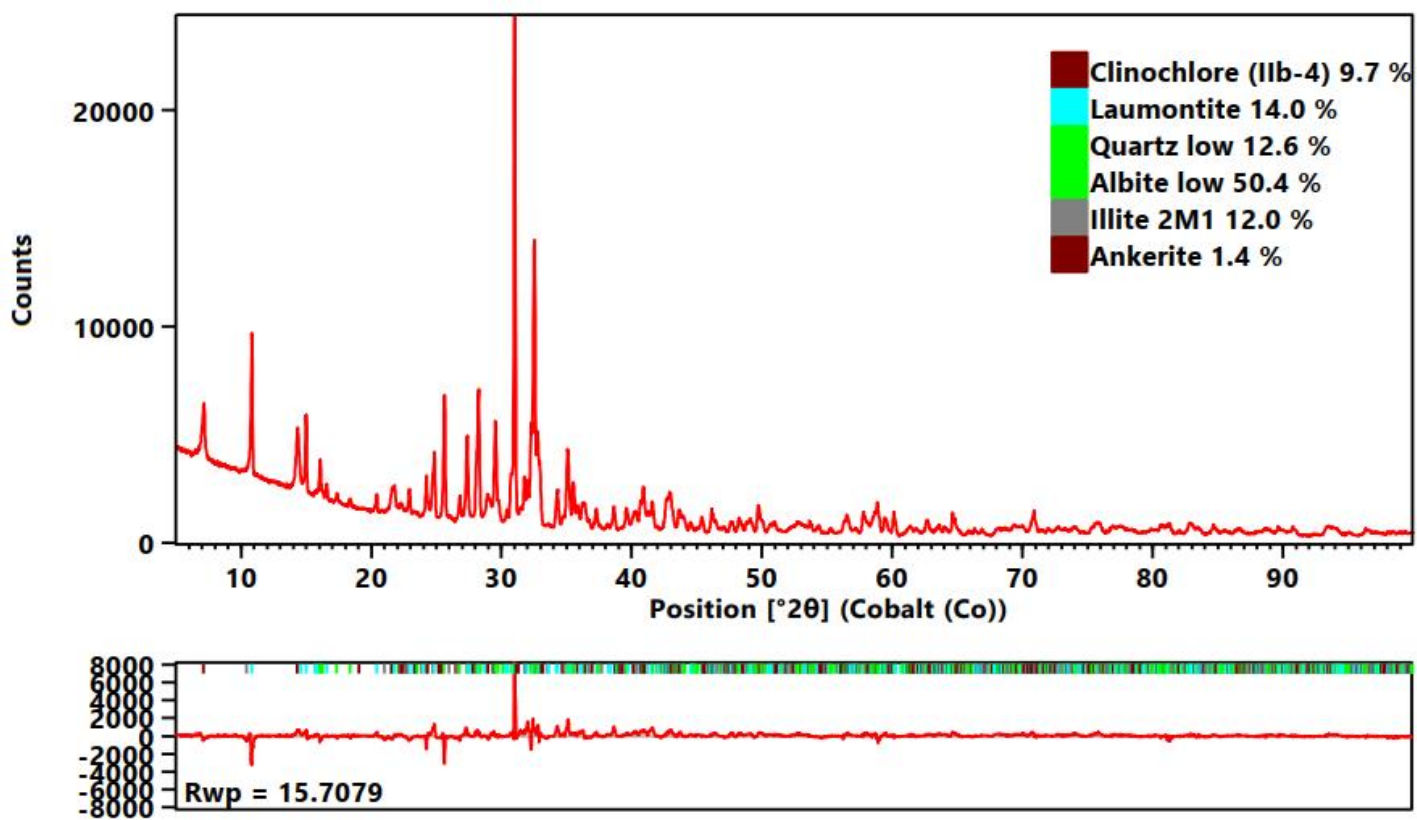

Figure 11. XRD from powder samples of the basaltic andesite sample and the difference plot from the Rietveld refinement with the associated mineral identification.

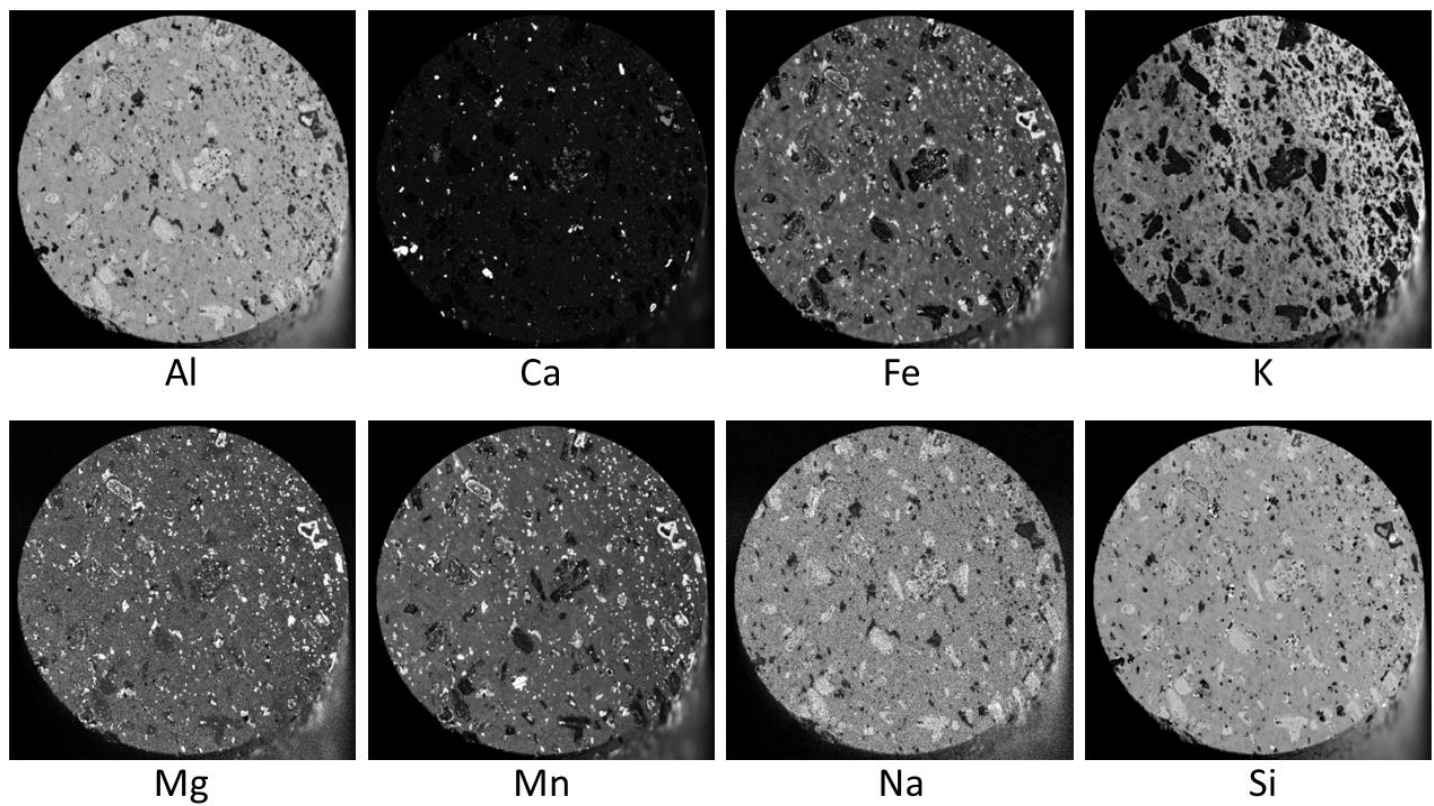

Figure 12. µXRF maps from the top surface of the training data shown with greyscale intensities.

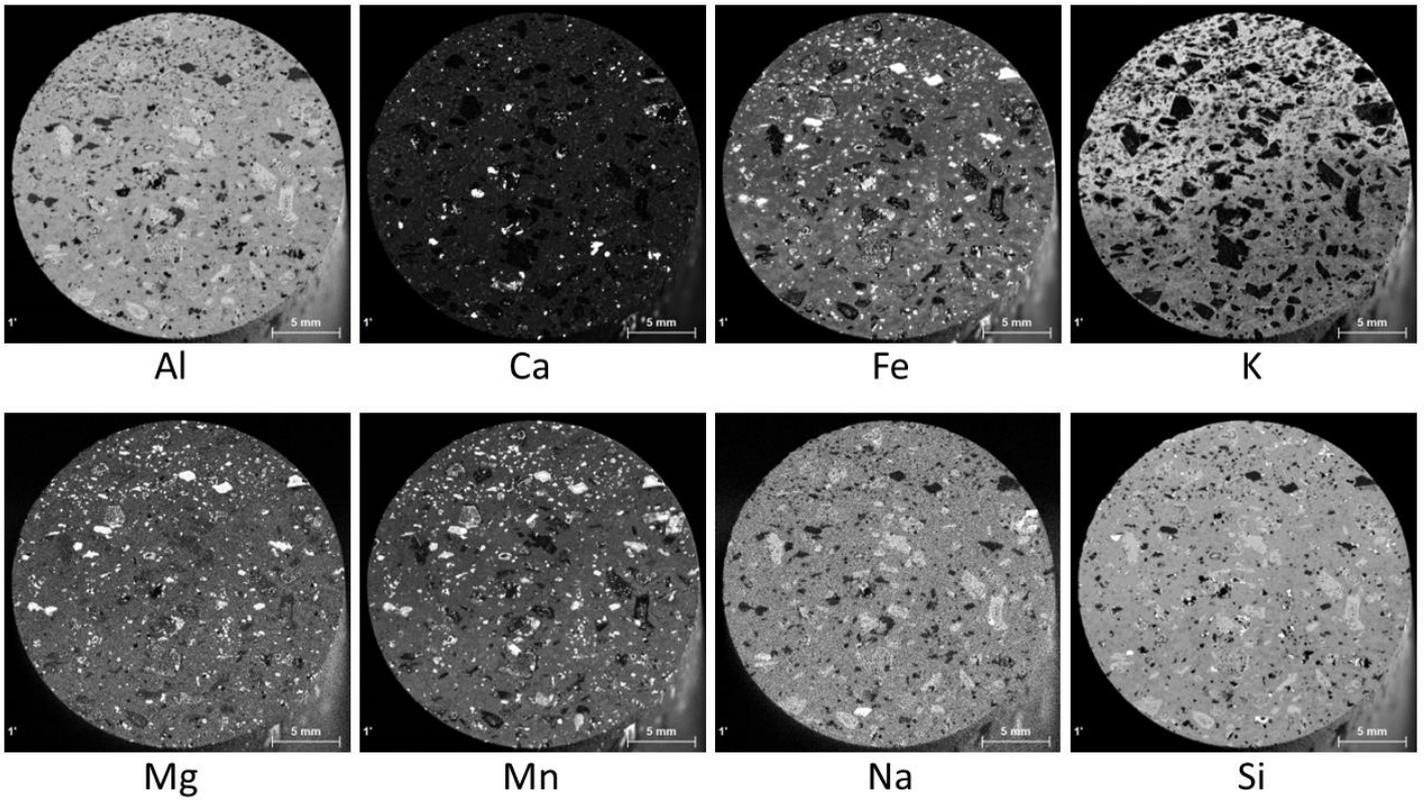

*Figure 13. µXRF maps from the bottom surface of the validation data shown with greyscale intensities.*

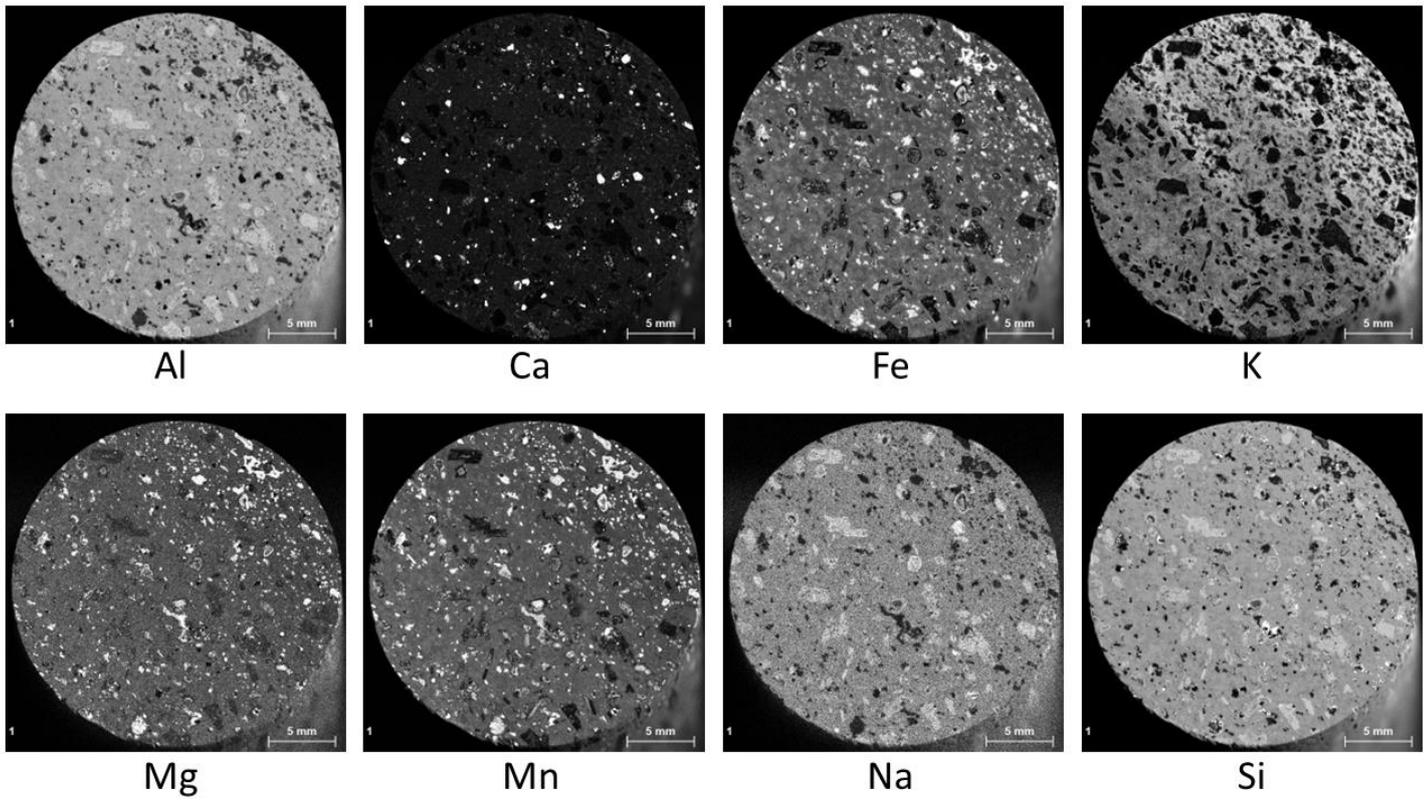

*Figure 14. µXRF maps from the top surface of the validation data shown with greyscale intensities.*

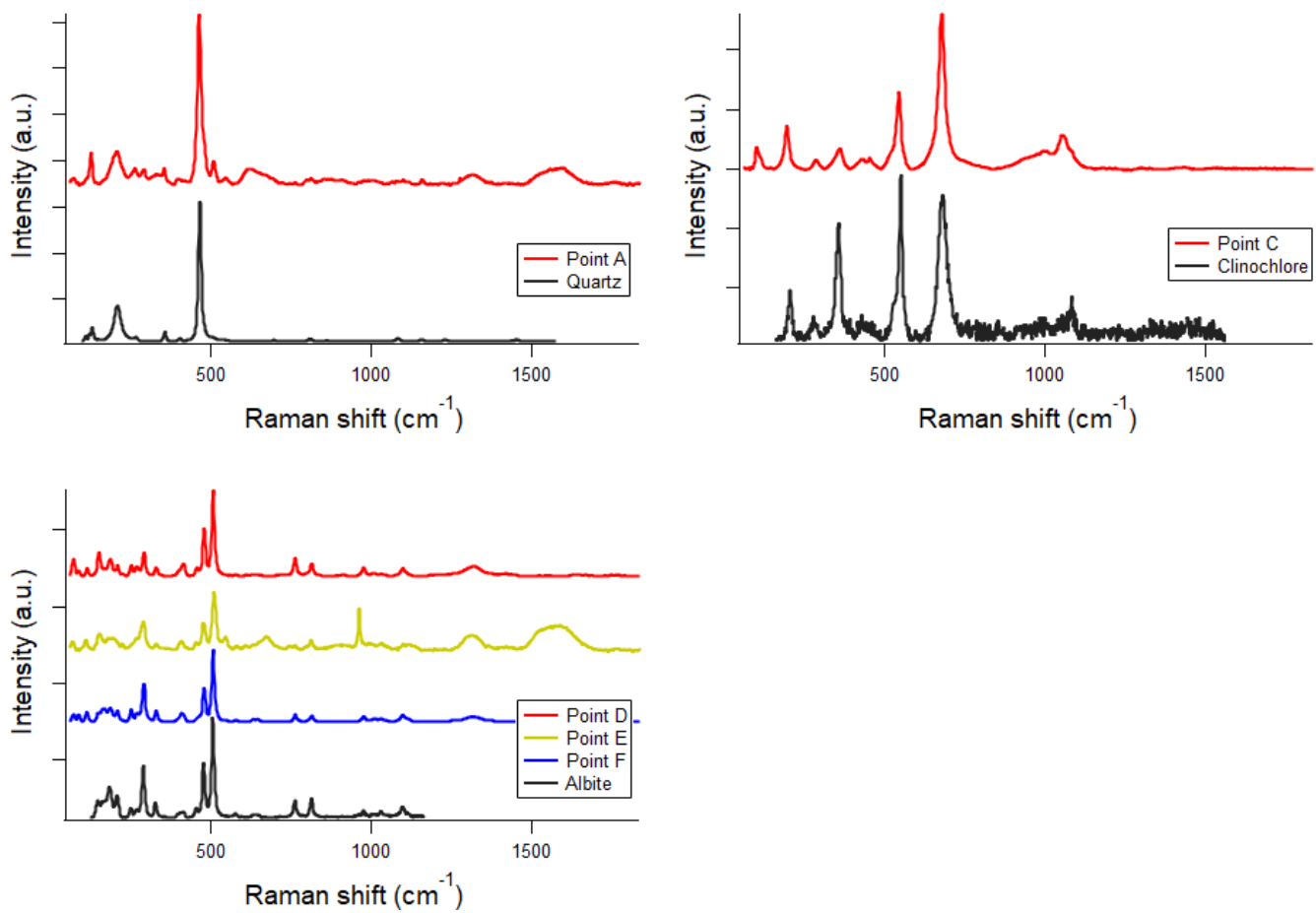

Figure 15. Raman spectra for quartz, clinochlore and albite shown in black plots with the spectra from their respective spots originating in Figure 2.